\documentclass[10pt,twocolumn,letterpaper]{article}

\usepackage[pagenumbers]{cvpr}      

\usepackage{xspace}
\usepackage{microtype}
\newcommand{\name}
{WorldReel\xspace}

\definecolor{cvprblue}{rgb}{0.21,0.49,0.74}
\usepackage[pagebackref,breaklinks,colorlinks,allcolors=cvprblue]{hyperref}

\def\paperID{} 
\def\confName{CVPR}
\def\confYear{2026}

\title{\name: 4D Video Generation with \\ Consistent Geometry and Motion Modeling}

\author{Shaoheng Fang$^1$ \quad Hanwen Jiang$^2$ \quad Yunpeng Bai$^1$ \quad Niloy J. Mitra$^{2,3}$ \quad Qixing Huang$^1$\\[5pt]
$^1$The University of Texas at Austin \quad $^2$Adobe Research \quad $^3$University College London\\
}

\begin{document}

\twocolumn[{%
\renewcommand\twocolumn[1][]{#1}%
\maketitle
\vspace*{-.35in}
\begin{center}
    \centering
    \captionsetup{type=figure}
    \includegraphics[width=1.0\linewidth]{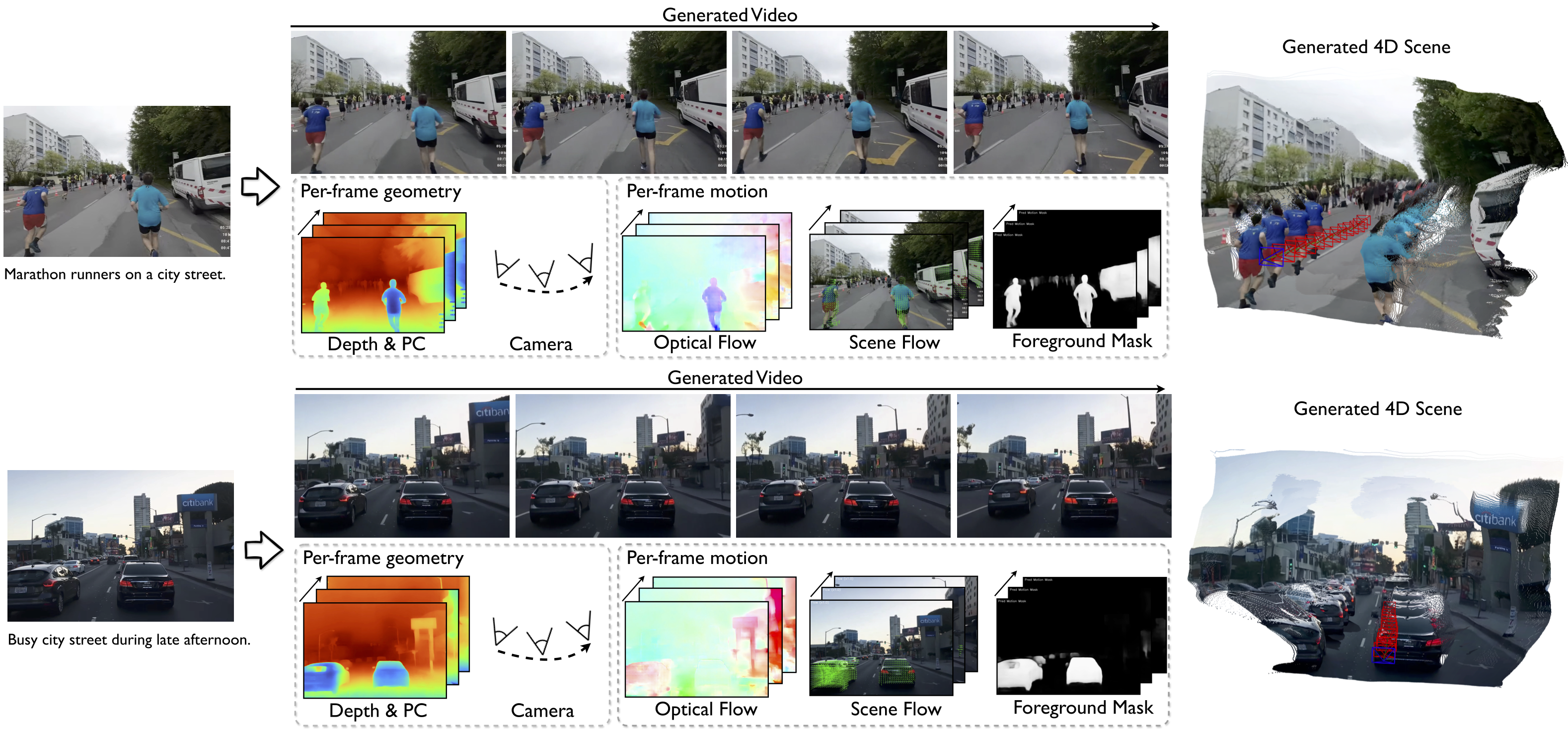}
    \captionof{figure}
{\textbf{End-to-end 4D generation.} 
Given a text prompt and a single input image (left), \name generates a video (center) \emph{together with} explicit 4D scene representations: per-frame geometry (depth + point cloud) with calibrated camera poses, and per-frame motion (optical flow, scene flow) with object masks (bottom panels). 
The rendered 4D scenes (right) exhibit consistent structure over time, even under non-rigid dynamics, illustrating spatio\-temporal consistency and tight coupling of appearance, geometry, and motion. Project page: \url{https://bshfang.github.io/worldreel/}
}
    \label{fig:fig1_overview}
\end{center}%
}]

\begin{abstract}
Recent video generators achieve striking photorealism, yet remain fundamentally inconsistent in 3D. We present \name, a 4D video generator that is natively spatio-temporally consistent. \name jointly produces RGB frames together with 4D scene representations, including pointmaps, camera trajectory, and dense flow mapping, enabling coherent geometry and appearance modeling over time. Our explicit 4D representation enforces a single underlying scene that persists across viewpoints and dynamic content, yielding videos that remain consistent even under large non-rigid motion and significant camera movement.
We train \name by carefully combining synthetic and real data: synthetic data providing precise 4D supervision (geometry, motion, and camera), while real videos contribute visual diversity and realism. This blend allows \name to generalize to in-the-wild footage while preserving strong geometric fidelity.
Extensive experiments demonstrate that \name sets a new state-of-the-art for consistent video generation with dynamic scenes and moving cameras, improving metrics of geometric consistency, motion coherence, and reducing view-time artifacts over competing methods~\cite{jiang2025geo4d,chen20254dnex,bai2025geovideo}. We believe that \name brings video generation closer to 4D-consistent world modeling, where agents can render, interact, and reason about scenes through a single and stable spatiotemporal representation.

\end{abstract}
\section{Introduction}

Recent video generators~\cite{wan2025wan, kong2024hunyuanvideo, yang2024cogvideox} have rapidly gained popularity, delivering impressive perceptual quality and realism across diverse prompts and scenes. Despite their impressive success in image fidelity and temporal smoothness, these models do \textit{not} maintain a single stable 3D scene that evolves over time. The resulting inconsistencies manifest as view-time drift, geometric flicker, and entangled camera/scene motion; these limitations become acute when extrapolating viewpoints or editing content as in emerging world model settings.


A natural next aspiration is 4D generation, in which a model maintains a coherent spatiotemporal scene representation. Prior efforts either optimize explicit dynamic scene representations~\cite{ren2023dreamgaussian4d, bah20244dfy, pan2024efficient4d} or post-hoc lift controllably generated 2D videos into 3D structure~\cite{zhao2024genxd, sun2024dimensionx, ren2025gen3c}. These approaches improve geometric alignment, but either are computationally heavy or fundamentally inherit the geometric inconsistency of 2D video priors, with limited capability to generalize to in-the-wild dynamics.
None natively integrates a true 4D structure into a generative prior.

We propose \textit{\name}, a unified 4D generator that jointly produces RGB, per-frame geometry, and motion in both 2D and 3D, explicitly outputting pointmaps, calibrated camera trajectories, and scene flow. In contrast to 2D optical flow or keypoint tracking used in prior methods~\cite{chefer2025videojam, jeong2025track4gen}, scene flow directly encodes 3D dynamics, (i) cleanly disentangling camera motion from object motion and (ii) operating in a physically-grounded evolving 3D frame. To prevent appearance from leaking into geometry/motion channels, we introduce an appearance-independent representation that stabilizes learning signals across viewpoints and lighting of both synthetic and real data.

At the core of \name is an augmented geometry-motion latent for a video diffusion transformer. This latent explicitly carries geometry and motion information through the generative process, yielding two benefits: (i) stronger inductive bias for 4D consistency and (ii) appearance-agnostic conditioning that improves generalization. Crucially, we demonstrate how to leverage this appearance-independent representation for effectively using synthetic datasets that provide accurate 4D labels (geometry, motion, camera) without sacrificing realism when mixed with real videos.

We design an architecture that predicts 4D from the geo-motion latent via multi-task learning with a customized temporal DPT-style decoder. Specifically, a lightweight shared backbone captures correlations among geometric tasks, with multiple task-specific heads for pointmaps, camera, dynamic mask, and scene flow. We train these outputs jointly with regularization terms that explicitly decouple static and dynamic components of the scene. This enforces geometric consistency on static structure and motion consistency on non-rigid regions, resulting in 4D coherence under both object and camera motion.

We extensively compare \name\ with state-of-the-art video and 4D generation models~\cite{chen20254dnex, bai2025geovideo, sun2024dimensionx} in terms of both (i)~geometry and motion consistency of generated videos, and (ii)~geometric quality of generated 4D scenes, where \name\ delivers consistent and substantial improvements. For video generation, \name\ produces stronger camera and subject motion, achieving the best dynamic degree on both general and complex motion, while maintaining state-of-the-art photorealism. For generated 4D scenes, our geometry is significantly more accurate, reducing depth error from $0.353 \rightarrow 0.287$ and achieving the lowest camera pose errors compared to recent 4D/3D-aware baselines~\cite{chen20254dnex,bai2025geovideo}. 

In summary, our main contributions are: 
(i)~\name as a unified generative framework that outputs RGB, pointmaps, calibrated cameras, optical flow, and scene flow, enforcing a persistent dynamic 3D scene through time; 
(ii)~an appearance-agnostic geo-motion latent that embeds explicit geometry and motion, improving generalization and enabling strong supervision from synthetic and real data; 
(iii)~a shared, lightweight DPT backbone with multi-task heads and targeted regularizers that decouple camera motion from dynamic components for tight geometric and motion consistency. 
As a result, \name sets a new SoTA in terms of 4D consistent output, especially for dynamic assets, taking a step towards 4D-consistent world modeling, where scenes can be rendered, edited, and reasoned about from a single stable spatiotemporal representation.

\section{Related Works}

\subsection{Diffusion Models for Video Generation}

The success of diffusion models in image synthesis~\cite{saharia2022photorealistic, hatamizadeh2024diffit} has been extended to video generation, demonstrating promising results~\cite{guo2023animatediff, ho2022video, singer2022make, wu2023tune}. Specifically, latent diffusion~\cite{rombach2022high, vahdat2021score, blattmann2023stable, chen2023videocrafter1, he2022latent} is applied, where a variational autoencoder~\cite{kingma2013auto} first encodes the image or video into a compact latent representation, and the diffusion process is performed in the latent space. This variational approach forms the basis for current state-of-the-art video generation models, including cascaded diffusion systems~\cite{ho2022imagen, wang2025lavie, zhang2023i2vgen} and large diffusion models~\cite{kong2024hunyuanvideo, hong2022cogvideo, yang2024cogvideox, wan2025wan} built on DiT backbones~\cite{peebles2023scalable}. However, a commonly recognized limitation of these models~\cite{hong2022cogvideo, blattmann2023stable} is their struggle with physical plausibility; they often fail to produce consistent 3D geometric structures and coherent temporal motions.

\noindent \textbf{Geometry-Aware Video Generation.} To address the lack of 3D consistency, many works have focused on integrating geometric priors into the generation process. Some methods jointly model video and geometry in the form of depth~\cite{huang2025voyager, dai2025fantasyworld, xi2025omnivdiff, bai2025geovideo} or point clouds~\cite{zhang2025world, dai2025fantasyworld, chen20254dnex}, enabling the simultaneous generation of video and its underlying geometric structure; 
OmniVDiff~\cite{xi2025omnivdiff} models appearance, depth, Canny edges, and semantic segmentation simultaneously, enabling multi-modal conditional control;  
GeoVideo~\cite{bai2025geovideo} additionally introduces an explicit geometric regularization loss, further improving 3D consistency in generated videos. Furthermore, \cite{wu2025geometry} uses a 3D foundation model~\cite{wang2025vggt} to implicitly incorporate geometric priors, aligning the video latent space with the foundation model’s features. However, these methods largely neglect scene dynamics and motion, focusing primarily on static scenes.

\noindent \textbf{Motion-Aware Video Generation.} In a parallel effort to improve temporal coherence, motion priors are incorporated into the generation process and have shown effective results. Track4Gen~\cite{jeong2025track4gen} adds an auxiliary point tracking task to provide explicit spatial supervision over diffusion latents. Additionally, optical flow, as a simple, generic, and easily obtainable motion representation, is frequently utilized. Videojam~\cite{chefer2025videojam} jointly models video and optical flow during generation, and introduces an inference-time guidance mechanism using the motion prediction as a dynamic guidance signal. In addition, optical flow is also useful as an input condition to achieve coherent and smooth motion during generation~\cite{shi2024motion, jin2025flovd}. Alternatively~\cite {nam2025optical, wang2025motif}, optical flow is used to assess motion quality and intensity, guiding video generation models to generate smoother/dynamic motions.

\subsection{Feed-Forward 4D Perception}

Latest 4D feed-forward models aim to recover camera parameters and consistent geometric properties, such as depth and tracking, directly from image sequences. 
Following the success of Dust3R~\cite{wang2024dust3r} on static scenes, several works have adapted the transformer-based framework to dynamic scenarios for image pairs~\cite{zhang2024monst3r, chen2025easi3r, han2025d} or longer sequences~\cite{wang2025continuous3dperceptionmodel, 11092390, zhuo2025streaming}.
Alternative approaches leverage large-scale foundation models. For instance, Geo4D~\cite{jiang2025geo4d} and GeometryCrafter~\cite{xu2025geometrycrafter} repurpose video generation models~\cite{xing2024dynamicrafter, blattmann2023stable} for multi-modal geometry generation, leveraging the strong dynamic priors learned by pre-trained video diffusion models.
Meanwhile, L4P~\cite{badki2025l4p} utilizes a pre-trained video encoder~\cite{wang2023videomae} to extract video features for various downstream 4D tasks.
In our method, we generate various 4D properties together with the videos.

\subsection{4D Generation}

Early methods~\cite{singer2023text4d, ren2023dreamgaussian4d, bah20244dfy, pan2024efficient4d} optimize explicit 4D scene representations~\cite{pumarola2020d, Wu_2024_CVPR} with score distillation sampling (SDS)~\cite{poole2022dreamfusion}, leveraging priors from video diffusion models~\cite{blattmann2023stable, singer2022make} and 3D‑aware multi‑view diffusion models~\cite{liu2023zero1to3, shi2023MVDream}.
These optimization-based pipelines suffer from high computational cost and long generation times, and are typically constrained to single dynamic object generation.
To improve scalability, recent methods have moved to feed‑forward prediction that outputs a 4D scene representation in a single pass. 
L4GM~\cite{ren2024l4gm} extends the LGM~\cite{tang2024lgm} to predict per‑frame 3D Gaussian splats, but remains limited to single‑object settings. 
Others~\cite{zhao2024genxd, sun2024dimensionx, ren2025gen3c, wu2025cat4d} leverage controllable video generation to synthesize dynamic videos, but require an additional reconstruction stage to create the 4D representation.
TesserAct~\cite{zhen2025tesseract} proposes jointly predicting RGB-DN (RGB, Depth, and Normal) videos to build a 4D world model. However, it is specifically tailored for embodied robotics and focuses largely on manipulation scenarios.
4DNeX~\cite{chen20254dnex} jointly models videos and point cloud geometry, training on dynamic datasets to enable dynamic point cloud generation. Despite this, their approach does not explicitly model scene dynamics and produces nearly fixed camera motion.

We follow these works to build our method on a video generation model. Moreover, we generate all 2D and 3D geometry and motion simultaneously from an augmented video latent space and design regularization terms to enforce motion and geometry consistency across time.

\section{\name{}}

\begin{figure*}[t]
    \centering
    \includegraphics[width=0.95\linewidth]{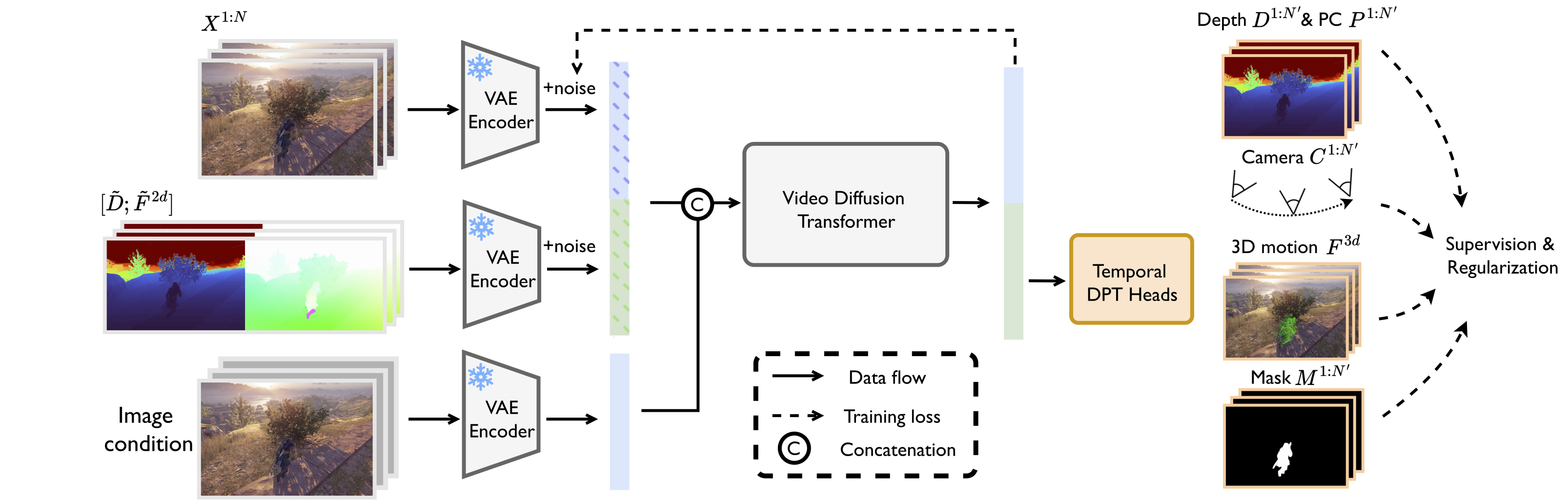}
    \caption{
    \textbf{Overview of \name.} We augment a video diffusion transformer with a geo–motion latent (from RGB and 2.5D cues such as depth/optical flow) to inject a 4D inductive bias for spatio-temporal consistency. A temporal DPT decoder is trained with direct supervision and regularization to predict unified 4D outputs (depth/point cloud, calibrated camera, 3D scene flow, and masks). 
    }
    \label{fig:fig2_architecture}
\end{figure*}

Based on existing powerful video generation models, we improve the spatio-temporal consistency of generated videos by introducing 4D inductive bias. In the following part, we first review the basics of video generation via latent diffusion (Sec.~\ref{sec:preliminary}). We then introduce two key designs of \name{}, including a latent space augmented by additional 2.5D inputs (Sec.~\ref{sec:latent_space}), and the unified 4D outputs that are directly supervised for improving consistent video and 4D geometry modeling (Sec.~\ref{sec:4d_modeling}).

\subsection{Preliminary: Video Latent Diffusion}
\label{sec:preliminary}
Our method is built on the latest latent diffusion video generation methods with transformer-based models~\cite{zheng2024open, wan2025wan, kong2024hunyuanvideo, yang2024cogvideox}, where the diffusion process occurs in a lower-dimensional latent space of a pre-trained 3D VAE~\cite{yang2024cogvideox}. Given a clean video clip $X^{1:N}\in\mathbb{R}^{N\times H\times W\times 3}$, the latent can be denoted as $\mathbf{z}_0 = \mathcal{E}(X^{1:N})$, where $\mathcal{E}$ is the encoder of the 3D VAE. The diffusion involves a forward process that adds a Gaussian noise $\epsilon\sim\mathcal{N}(0,I)$ to obtain the noisy latent $\mathbf{z}_t=\alpha_t \mathbf{z}_0+\sigma_t \epsilon$, where $t=1, \cdots, T$ is the diffusion time step and $\alpha_t$ and $\sigma_t$ are noise scheduler parameters. The reverse process is to remove noise through a denoiser $f_\theta$ with condition $\mathbf{c}$, trained by minimizing:
\begin{equation}
    \min_{\theta}\;
\mathbb{E}_{t\sim{U}[1,T],\,\epsilon\sim\mathcal{N}(0,I)}
\left[ \bigl\| f_\theta(\mathbf{z}_t,t,\mathbf{c})-\epsilon \bigr\|_2^2 \right]
\label{eq:diffusion_objective}
\end{equation}


\subsection{Geometry-Motion Augmented Latent}
\label{sec:latent_space}
Our objective is to generate dynamic 4D scenes using a video diffusion model, which requires the ability to jointly model video appearance, 3D geometry, and complex motion. To achieve this, we propose to extend the latent space of video generation models by introducing explicit priors for geometry and motion. We use frame-aligned depth maps and optical flow to jointly model 2.5-D geometry and motion. Concretely, depth captures the scene structure under perspective, while optical flow summarizes the per-pixel displacement induced by both camera and object motion.

Our choice of depth and optical flow as the preliminary geo-motion representation is based on several advantages: i) Both depth maps and optical flow fields possess a dense, image-like structure that is intrinsically aligned to RGB frames. This modality compatibility allows us to leverage powerful, pre-trained 3D VAEs to directly encode and decode them (as shown in \cite{ke2024repurposing, hu2025depthcrafter}) and is easy to integrate into existing video generation pipelines. ii) High-quality depth maps and optical flow are readily obtainable from powerful foundation models~\cite{hu2025depthcrafter, wang2024sea}, which enables our model training to be conducted at scale. iii) The geo-motion latent is 3D-focused, which factors out the appearance and texture of the generated scene, resulting in a smaller distribution gap between synthetic and real-world data. When training subsequent 4D generation tasks, this disentanglement enables better leverage of accurate ground-truth annotations from synthetic datasets, leading to stronger generalization performance.

To encode the additional inputs of depth $D_i \in \mathbb{R}^{H\times W\times 1}$, and forward optical flow $F^{\text{2d}}_i \in \mathbb{R}^{H\times W\times 2}$ with existing 3D VAE, we need to align them to the same value range with the RGB images.
Thus, we get the normalized depth and optical flow as $\tilde{D_i} = 2 \cdot \frac{D_i - d_{min}}{d_{max} - d_{min}} - 1$ and $\tilde{F}^{\text{2d}}_i = \frac{F^{\text{2d}}_i}{| F^{\text{2d}} |_{max}}$, where $d_{max}$ and $d_{min}$ are the maximum and minimum depth and $| F^{2d} |_{max}$ is the maximum displacement scale across all frames.
The geo-motion latent can be encoded by the pre-trained 3D VAE, denoted as $\mathbf{z}^{\text{gm}}_0 = \mathcal{E}([\tilde{D}; \tilde{F}^{\text{2d}}])$, where $[\cdot; \cdot]$ denotes concatenation along the channel dimension. With the original video latent $\mathbf{z}^{\text{rgb}}_0 = \mathcal{E}(X)$, we can obtain the augmented latent for the diffusion model by channel-wise concatenation $\mathbf{z}_0 = [\mathbf{z}^{\text{rgb}}_0 ; \mathbf{z}^{\text{gm}}_0]$

\noindent \textbf{Adapt Existing Video Generation Models.} We then train the video diffusion DiT to model the joint distribution over appearance and geo-motion in the augmented latent space. To effectively leverage pre-trained weights, we adapt the original transformer architecture with minimal modifications. Specifically, only the input and output projection layers are modified for the doubled channel dimension of $\mathbf{z}_0$, while all intermediate blocks remain unchanged. To further stabilize training, we employ a zero-initialization scheme for the input projection layer: weights corresponding to the original video latent $\mathbf{z}^{rgb}$ are loaded from the pre-trained model, while new expanded parameters corresponding to the geo-motion latent $\mathbf{z}^{gm}$ are initialized to zero. This strategy ensures that, at the beginning of training, the model's behavior is identical to that of the original video diffusion model, improving training stability.

\subsection{4D Video Modeling}
\label{sec:4d_modeling}

\noindent\textbf{Unified 4D Representations as Outputs.} 
While leveraging additional inputs of depth and optical flow provides strong priors for improving video generation in latent space, they are insufficient to recover the 3D scene structure, especially for disentangling camera and object motion which is ill-posed under the 2.5-D setting. Thus, we opt to predict finer-grained 4D representations of the generated video clip and apply explicit supervision. In detail, except for the video clip, the model also outputs the corresponding camera trajectory, point clouds, and object foreground masks. The supervision of 4D representations enables consistent spatial alignment across frames, by back-propagating geometry-related gradients to the latent space. This design choice also effectively helps the disentanglement between camera motion and object motion, encouraging the model to capture 3D dynamics in a better latent space.


Formally, for a downsampled set of latent frames $i=1, \cdots N^{\prime}$, we define the output unified 4D representation as $( D_i, P_i, C_i, F_i^{\text{3d}}, M_i )$. $C_i \in \mathbb{R}^9$ are camera intrinsics and extrinsics following the parameterization in \cite{wang2025vggt}. $P_i \in \mathbb{R}^{H\times W \times 3}$ is the point cloud, $F_i^{3d} \in \mathbb{R}^{H\times W \times 3}$ is the forward scene flow, and $M_i \in \mathbb{R}^{H \times W}$ is the dynamic (foreground) mask. Camera parameters $C_i$, point clouds $P_i$, and scene flow $F_i^{\text{3d}}$ are all represented in the first‑frame canonical coordinate system for a consistent scene description. 

\noindent\textbf{Model Design.}
To predict the unified 4D representations, we design a customized temporal DPT~\cite{ranftl2021vision} architecture. The model first extracts multi-scale dense features from the input latent, which are then processed and aggregated using a DPT-like fusion backbone incorporating temporal transformers (see supp.). Critically, our design is motivated by the high correlation between target geometric representations. We utilize a single, shared DPT-style decoder to process and generate unified dense features. Only at the final output layer do multiple lightweight, task-specific heads predict their respective tasks. This not only ensures significant parameter efficiency but also acts as a strong form of regularization, forcing the model to learn a unified and geometrically consistent representation for all tasks.

\noindent\textbf{Training.} Our model is trained using a two-stage strategy to ensure stability and optimize both generative quality and 4D accuracy. The first stage trains the video diffusion transformer and the temporal DPT heads in isolation. The second stage then jointly trains the full model end-to-end with additional regularization terms. 

Specifically, we first finetune the video diffusion transformer to accommodate the new geo-motion augmented latent. We use a standard diffusion loss as defined in Eq.~\ref{eq:diffusion_objective}. The total loss combines losses on the appearance component and the geo-motion components: $\mathcal{L}_{\text{diff}} = \mathcal{L}_{\text{diff}}^{\text{rgb}} + \mathcal{L}_{\text{diff}}^{\text{gm}}$.

To pre-train the temporal DPT heads on 4D prediction, we use the clean geo-motion latent $\mathbf{z}_{0}^{\text{gm}}$ as input and optimize using a comprehensive multi-task loss $\mathcal{L}_{\text{dpt}}$.
\begin{equation}
    \mathcal{L}_{\text{dpt}} =\mathcal{L}_{\text{depth}} + \mathcal{L}_{\text{pc}} + \mathcal{L}_{\text{cam}} + \mathcal{L}_{\text{mask}} + \lambda_{\text{flow}}\mathcal{L}_{\text{flow}}.
    \label{eq:dpt_loss}
\end{equation}
We use masked L1 loss for $\mathcal{L}_{\text{depth}}$ and $\mathcal{L}_{\text{pc}}$ on pixels with valid depth values; Huber loss for $\mathcal{L}_{\text{cam}}$; BCE loss for $\mathcal{L}_{\text{mask}}$. In $\mathcal{L}_{\text{flow}}$, we reweight the loss pixel-wise to focus on the prediction of the foreground motion according to the dynamic mask $\hat{M}_i$ (see supp. for details).

Then, we jointly train the video transformer model and the temporal DPT heads end-to-end. Regularization terms are added for geometry and temporal consistency. To ensure geometry consistency in the static background and motion smoothness for dynamic objects, we apply different regularization terms. We use $\hat{M}^{\text{bg}}_{i}$ and $\hat{M}^{\text{fg}}_{i}$ to represent the background and foreground masks derived from $\hat{M}_i$ label. 
\begin{equation}
\begin{gathered}
  \mathcal{L}_{\text{reg}}^{\text{depth}} = \sum_{i} \sum_{j} \left\|  \hat{M}^{\text{bg}}_{i} \odot \left( D_{j} - \text{Proj}(D_i, T_{i \to j}) \right) \right\|_2 \\
  \mathcal{L}_{\text{reg}}^{\text{flow}} = \sum_i \left( \left\| \hat{M}_i^{\text{fg}} \odot \nabla_x F_{i}^{\text{3d}} \right\|_2 + \left\| \hat{M}_i^{\text{fg}} \odot \nabla_y F_{i}^{\text{3d}} \right\|_2 \right) \\
  \mathcal{L}_{\text{reg}} = \mathcal{L}_{\text{reg}}^{\text{depth}} + \mathcal{L}_{\text{reg}}^{\text{flow}}
\end{gathered}
\end{equation}

Here $T_{i \to j}$ is the transformation matrix derived from $C_i, C_j$. The joint training objective is

\begin{equation}
    \mathcal{L} = \mathcal{L}_{\text{diff}} + \lambda_{\text{dpt}} \mathcal{L}_{\text{dpt}} + \lambda_{\text{reg}} \mathcal{L}_{\text{reg}}
    \label{eq:joint_loss}
\end{equation}

\subsection{4D Data Preparation}
\label{sec:4d_data_prep}

Training a model for comprehensive 4D video generation requires a diverse, high-quality video dataset with corresponding 4D labels, encompassing a wide range of dynamics and sufficient scene complexity. Therefore, we employ a mixed-data strategy~\cite{zheng2023pointodyssey, black2023bedlam, karaev2023dynamicstereo, zhou2025omniworld, wang2025spatialvidlargescalevideodataset}, leveraging both synthetic datasets with ground-truth labels and real-world videos with pseudo-annotations.

Synthetic datasets include PointOdyssey~\cite{zheng2023pointodyssey}, BEDLAM~\cite{black2023bedlam}, Dynamic Replica~\cite{karaev2023dynamicstereo}, and Omniworld‑Game~\cite{zhou2025omniworld}. 
Synthetic data alone is limited in scale and complexity, which can impede the model's generalization to real-world scenarios. We therefore supplement our training with real-world videos.
While some recent works construct 4D datasets~\cite{wang2025spatialvidlargescalevideodataset, chen20254dnex} by filtering large-scale sources~\cite{chen2024panda, fan2025vchitect} and using off-the-shelf annotation models~\cite{zhang2024monst3r, li2025megasam}, the resulting labels are noisy and of insufficient quality. We start with high-quality raw videos filtered from Panda-70M~\cite{chen2024panda} by SpatialVid~\cite{wang2025spatialvidlargescalevideodataset} and re-annotate them using state-of-the-art vision foundation models to obtain superior pseudo-labels. Specifically, for per-frame depth, we use GeometryCrafter~\cite{xu2025geometrycrafter} to obtain temporally smooth depth sequences. For feed-forward 4D task labels corresponding to latent frames, we employ ViPE~\cite{huang2025vipe} to obtain camera parameters, depth maps, and foreground masks. We obtain the point cloud annotations by back-projecting the depth map with camera poses, and all camera poses and point clouds are in the first-frame canonical coordinates.

\noindent\textbf{2D/3D Motion Labels.} Accurate motion supervision is critical, yet ground‑truth scene flow is rarely available. For all datasets, we estimate frame-to-frame optical flow using SEA-RAFT~\cite {wang2024sea}. Then, inspired by zero-MSF~\cite{liang2025zero}, we compute large‑scale, dense scene flow labels from optical flow and corresponding geometry labels.
We derive dense 3D scene-flow pseudo-labels $\hat{F}^{\text{3d}}_i \in \mathbb{R}^{H\times W\times 3}$ in the first-camera canonical frame for each adjacent pair $(X_i, X_{i+1})$.
We first get the forward and backward optical flow $F^{\text{2d}}_{i\rightarrow i+1}, F^{\text{2d}}_{i+1\rightarrow i} \in\mathbb{R}^{H\times W\times 2}$, and their per-pixel uncertainties $\sigma^{\text{2d}}_{i\rightarrow i+1},\ \sigma^{\text{2d}}_{i+1\rightarrow i}\in\mathbb{R}^{H\times W}$ using SEA-RAFT~\cite{wang2024sea}.
For pixel $\mathbf{u} = (x,y)$ in frame $i$, define the forward-mapped pixel $\mathbf{q}(\mathbf{u}) = \mathbf{u} + F^{\text{2d}}_{i\rightarrow i+1}(\mathbf{u})$. Scene flow is then retrieved by
\begin{equation}
{\hat{F}}^{\text{3d}}_i(\mathbf{u}) = 
\begin{cases}
  P_{i+1}(\mathbf{q}(\mathbf{u}))-P_i(\mathbf{u}), & \text{if } \hat{M}_i(\mathbf{u}) = 1, \\
  \mathbf{0}, & \text{otherwise.}
\end{cases}
\end{equation}

\begin{table*}[t]
\small
\centering
\setlength{\tabcolsep}{4.5pt}
\caption{Quantitative comparison on image-to-video (I2V) generation under two splits: \emph{General motion} and \emph{Complex motion}. Metrics: dynamic degree (d.d.; $\uparrow$), motion smoothness (m.s.; $\uparrow$), I2V-subject/background (i2v-s./i2v-b.; $\uparrow$), subject consistency (s.c.; $\uparrow$), Fréchet Video Distance (FVD; $\downarrow$), and FID ($\downarrow$). \name achieves the best overall performance, with notably higher dynamic degree while maintaining strong s.c. and perceptual quality (lower FVD/FID). \textbf{Bold} indicates best; \underline{underline} second-best. \textcolor{gray}{Gray} rows denote methods that primarily focus on nearly-static scenes. 
}
\vspace{-0.05in}
\begin{tabular}{r|ccccc|cc|ccccc|cc}
\toprule
           & \multicolumn{7}{|c|}{General motion}                                                                                  & \multicolumn{7}{c}{Complex motion}                                                                                 \\
\hline
method     & d.d.           & m.s.           & i2v-s.         & i2v-b.         & s.c.           & FVD            & FID            & d.d          & m.s.           & i2v-s.         & i2v-b.         & s.c.           & FVD            & FID            \\
\hline
\hline
Cogvideox-I2V~\cite{yang2024cogvideox}  & 0.37          & 0.976          & 0.967          & 0.970          & 0.928          & 617.4          & 47.69          & 0.52         & 0.972          & 0.954          & 0.960          & 0.916          & 824.8          & 52.97          \\
4DNeX~\cite{chen20254dnex}      & {\color{gray} 0.03}          & {\color{gray} \textbf{0.994}} & {\color{gray} \textbf{0.993}} & {\color{gray} \textbf{0.990}} & {\color{gray} \textbf{0.983}} & {\color{gray} 712.5}          & {\color{gray} 44.97}          & {\color{gray} 0.19}         & {\color{gray} \textbf{0.994}} & {\color{gray} \textbf{0.987}} & {\color{gray} \textbf{0.985}} & {\color{gray} \textbf{0.983}} & {\color{gray} 632.8}          & {\color{gray} 49.79}          \\
DimensionX~\cite{sun2024dimensionx} & 0.47          & 0.987          & 0.974          & 0.979          & 0.943          & 470.7          & \underline{42.28}    & \underline{0.93}   & 0.980          & 0.963          & 0.970          & 0.910          & 605.3          & \underline{46.97}    \\
GeoVideo~\cite{bai2025geovideo}   & \underline{0.54}    & 0.987          & 0.979          & 0.980          & 0.932          & \underline{371.3}    & 46.78          & 0.79         & 0.985          & 0.971          & 0.974          & 0.914          & \underline{409.9}    & 49.92          \\
\hline
\name (ours)       & \textbf{0.73} & \underline{0.990}    & \underline{0.986}    & \underline{0.986}    & \underline{0.953}    & \textbf{336.1} & \textbf{36.58} & \textbf{1.00} & \underline{0.987}    & \underline{0.980}    & \underline{0.980}    & \underline{0.927}    & \textbf{394.2} & \textbf{44.95} \\
\bottomrule
\end{tabular}

\label{table:videogen}
\end{table*}

The scene flow labels generated by finding the correspondence in adjacent point clouds may contain large noise. Therefore, we utilize various existing information to eliminate potentially incorrect scene flow labels. A per-pixel validity mask $M_i^{\text{flow}}$ only keeps pixels that pass foreground/instance, uncertainty, and forward-backward consistency checks (see supp.)
During training, the motion validity mask is applied when calculating $\mathcal{L}_{\text{flow}}$ and $\mathcal{L}_{\text{reg}}^{\text{flow}}$. Also, to align the frame number for 2D/3D flow, we retrieve one more frame after the video clip during all data processing.

\section{Experiments}

\subsection{Experimental Settings}

\noindent\textbf{Implementation Details.} We utilize CogVideoX-5B-I2V~\cite{yang2024cogvideox} as our base video generation model, which generates videos at a resolution of $480\times720$ over $49$ frames. Our predicted 4D dynamic scene representations are generated at the same resolution but for a downsampled set of $13$ frames.
All training is conducted on the same mixed dataset of synthetic and real videos described in Sec.~\ref{sec:4d_data_prep}, which also details the generation process for pseudo annotations. 
Following the two-stage strategy described in Sec.~\ref{sec:4d_modeling}, we first finetune the geo-motion augmented DiT for 20K steps and separately train the temporal DPT heads from scratch for 100K steps. In the second stage, the entire model is trained end-to-end for an additional 10K steps.
All training is performed on 8$\times$H200 GPUs with a batch size of 8. We employ the AdamW optimizer~\cite{loshchilov2017fixing} with a constant learning rate of 2e-5. 
For optimization objectives, we set the loss weight $\lambda_{\text{flow}} = 5.0$ in Eq.~\ref{eq:dpt_loss} and the final joint loss weights $\lambda_{\text{dpt}}=0.1$ and $\lambda_{\text{reg}}=0.5$ in Eq.~\ref{eq:joint_loss}.

\noindent\textbf{Evaluation.}
To assess 4D generation for dynamic scenes, we construct two benchmarks from the SpatialVid~\cite{wang2025spatialvidlargescalevideodataset} validation split. The (i)~\emph{general} motion set contains 500 randomly sampled videos, while the (ii)~\emph{complex} motion set consists of 500 videos with the highest 3D motion magnitude, representing scenes with the most complex dynamics, ensuring a rigorous test of dynamic scene modeling (see supp. for details). We compare \name with the base model CogVideoX‑I2V~\cite{yang2024cogvideox} and other recent 4D video generation methods: DimensionX~\cite{sun2024dimensionx}, 4DNeX~\cite{chen20254dnex}, and GeoVideo~\cite{bai2025geovideo}. We use the released checkpoints for DimensionX~\cite{sun2024dimensionx} and 4DNeX~\cite{chen20254dnex}, and train GeoVideo~\cite{bai2025geovideo} on our dataset. For dynamic-scene video generation quality, we report five metrics from VBench~\cite{huang2024vbench}: (a)~i2v‑subject, (b)~i2v‑background, (c)~subject consistency, (d)~motion smoothness, and (e)~dynamic degree, which jointly measure the quality and consistency of dynamics in a video. We also report FVD~\cite{unterthiner2018towards} and FID~\cite{heusel2017gans} for video visual quality.

For \textit{4D geometry generation} assessment, we evaluate the geometry terms generated with the videos. We utilize ViPE~\cite{huang2025vipe}, the state-of-the-art dynamic-scene reconstruction method, to obtain a pseudo ground truth aligned with the video. We normalize the scene's scale using the median depth. For generated depth, we report log-RMSE and $\delta$-accuracy. For generated camera trajectories, we report the camera motion magnitude using the trajectory length (sum of inter-frame translations) and the cumulative viewpoint change (sum of inter-frame rotation angles). In addition, we use standard error metrics to evaluate the camera poses generated: Absolute Translation Error (ATE), Relative Translation Error (RTE), and Relative Rotation Error (RRE).

\begin{figure*}[t]
    \centering
    \includegraphics[width=0.98\linewidth]{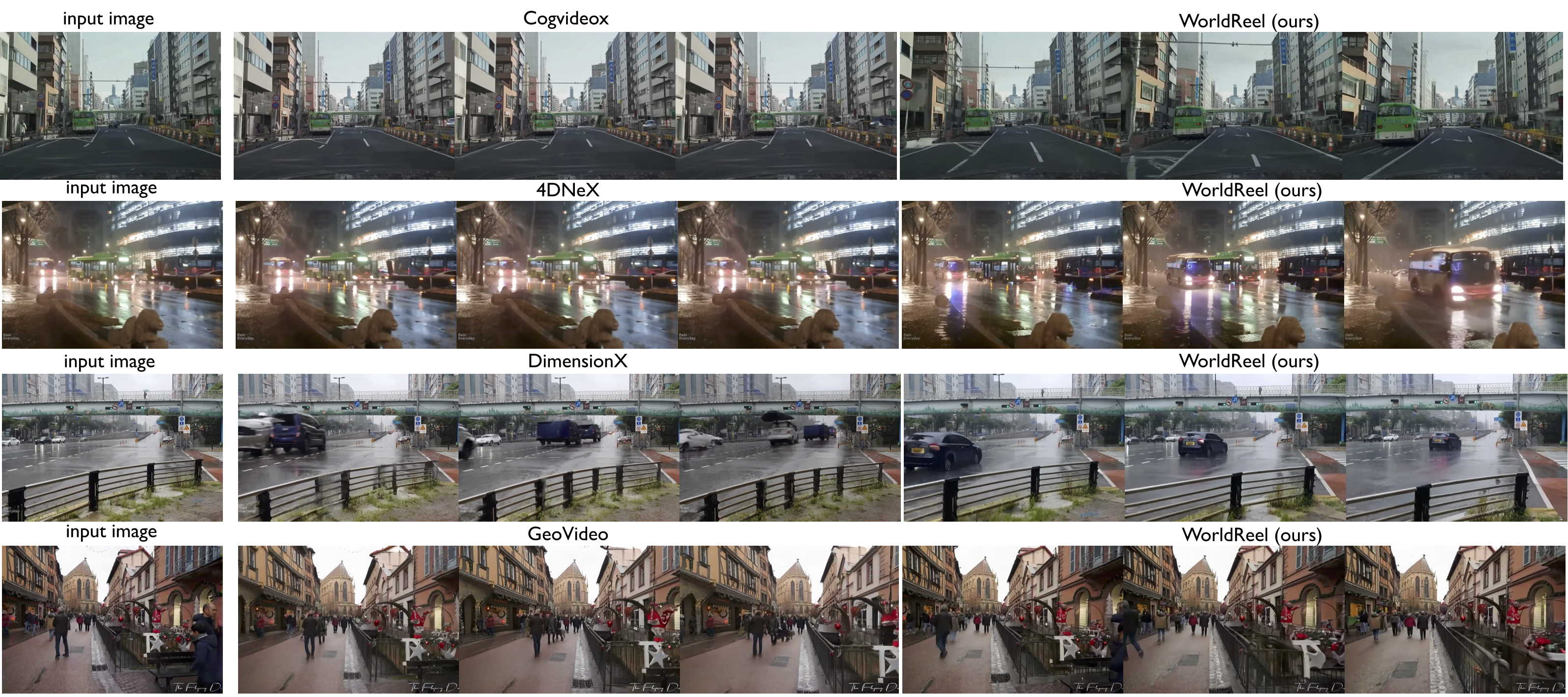}
    \vspace{-2mm}
    \caption{Qualitative image-to-video comparison on in-the-wild scenes. Given a single \emph{input image} (left), we show sampled frames from videos generated by 4DNeX~\cite{chen20254dnex}, DimensionX~\cite{sun2024dimensionx}, GeoVideo~\cite{bai2025geovideo}, and \emph{\name} (ours). Prior methods often exhibit geometry drift and motion inconsistencies (e.g., warped facades, misaligned vehicles), while our results better preserve scene layout and maintain coherent camera and non-rigid dynamics. See the supplementary for prompts, full videos for all methods, and additional comparisons.}
    \label{fig:i2v_result_baselines}
    \vspace*{-0.1in}
\end{figure*}

\subsection{Experimental Results}

\noindent\textbf{Video Generation Comparisons.} Table~\ref{table:videogen} demonstrates the quantitative comparison with other baseline models. Across both the \emph{general} and \emph{complex} motion sets, our method shows superior performance. 
Our model substantially improves visual quality: 
relative to GeoVideo~\cite{bai2025geovideo} that is trained on the same data, FVD drops from $\text{371.3}$ to $\textbf{336.1}\,(\text{-9.5\%})$ on \emph{general} and from $409.9$ to $\textbf{394.2}\,(\text{-3.8\%})$ on \emph{complex}.
Most notably, our model shows a significant improvement in the dynamic degree metric, far exceeding all baselines and achieving a perfect $1.0$ on the \emph{complex} motion set. 
Furthermore, our method achieves leading scores on motion smoothness, i2v-subject, i2v-background, and subject consistency, indicating our ability to generate smoother and more consistent dynamics.
These results demonstrate that our method significantly enhances the effective motion of both the camera and the scene \textit{without} sacrificing appearance quality or temporal consistency (see \Cref{fig:i2v_result_baselines}).

While GeoVideo~\cite{bai2025geovideo} introduces geometry modeling and regularization for consistency, our results suggest that the focus on static geometry penalizes the generation of dynamic content. In contrast, \name, by jointly and explicitly modeling both geometry and motion, avoids this trade-off and prevents the model from preferring static content to maintain geometric consistency. 
Also, although 4DNeX~\cite{chen20254dnex} reports high framewise consistency (e.g., s.c.), its extremely low dynamic degree ($0.03$) and poor FVD ($712.5$) indicate a collapse toward near‑static video generation.

\begin{table}[!t]
\footnotesize
\centering
\setlength{\tabcolsep}{3.5pt}
\caption{Scene geometry evaluation on depth, camera pose, and camera trajectory. \name achieves the best depth and camera accuracy across all pose metrics, with competitive trajectory estimates. Ablations (\emph{w/o geomotion}, \emph{w/o joint}) degrade depth or pose quality, confirming the importance of geo–motion latent and joint training. \textbf{Bold} = best, \underline{underline} = second-best.}
\vspace{-0.05in}
\resizebox{\linewidth}{!}{
\begin{tabular}{r|cc|ccc|cc}
\toprule
                        & \multicolumn{2}{|c|}{Depth}         & \multicolumn{3}{|c|}{Camera}                              & \multicolumn{2}{c}{Camera traj.} \\
\hline
method                  & log-rmse        & $\delta_{1.25}$       & ATE               & RTE               & RRE             & length      & rotation       \\
\hline
4DNeX~\cite{chen20254dnex}                   & 0.479          & 39.9          & \underline{0.006}    & 0.017          & 0.378          & 0.034           & 0.55         \\
GeoVideo~\cite{bai2025geovideo}                & 0.353          & 63.4          & 0.011          & \underline{0.012}    & 0.443          & 0.331           & 4.61          \\
\hline
w/o geomotion      & \underline{0.352}    & \underline{67.2}    & 0.010          & 0.013          & \underline{0.372}    & \textbf{0.379}  & \underline{5.34}    \\
w/o joint & 0.399          & 57.6          & 0.006          & 0.014          & 0.410          & 0.294           & \textbf{5.86} \\
\name                    & \textbf{0.287} & \textbf{71.1} & \textbf{0.005} & \textbf{0.007} & \textbf{0.317} & \underline{0.358}     & 3.83       \\
\bottomrule
\end{tabular}
}
\label{table:scene_geometry}
\vspace{-0.2in}
\end{table}

\noindent\textbf{4D Scene Quality.} \Cref{table:scene_geometry} assesses the fidelity of the generated 4D scene geometry. Because 4DNeX~\cite{chen20254dnex} outputs raw point clouds, we follow their pipeline to register frames and recover camera poses. Although 4DNeX attains a low camera ATE, its trajectory length and rotation are near-zero, indicating little actual camera motion, consistent with its tendency to collapse toward static scenes observed in our video metrics. In contrast, \name delivers the best depth and pose accuracy across all measures, with camera parameters that faithfully match the generated videos.

Ablations further underscore our design choices: removing the joint training stage (“w/o joint”) noticeably degrades both geometric and camera accuracy. This confirms that our joint optimization with targeted regularization -- decoupling static structure from dynamic regions and supervising them separately -- is crucial for high 4D consistency. The resulting high-quality geometry and poses highlight WorldReel's potential for realistic dynamic 4D scene generation.

\begin{table*}[t]
\small
\centering
\setlength{\tabcolsep}{4.5pt}
\caption{Ablation study on image-to-video  generation under \emph{General} and \emph{Complex} motion. Variants: \emph{base finetuned}, \emph{w/o g.m.} (without geo–motion latent), \emph{w/o joint} (no joint multi-task decoding/regularizers), \emph{freeze dpt} (freeze temporal DPT backbone in stage-2), and \emph{full} (ours). Metrics: dynamic degree (d.d.; $\uparrow$), motion smoothness (m.s.; $\uparrow$), I2V-subject/background (i2v-s./i2v-b.; $\uparrow$), subject consistency (s.c.; $\uparrow$), FVD ($\downarrow$), and FID ($\downarrow$). Our \emph{full} model delivers the best overall quality (lowest FID and highest d.d. on complex motion), while \emph{freeze dpt} attains the lowest FVD; removing geo–motion latent or joint training degrades consistency. \textbf{Bold} = best, \underline{underline} = second-best.}
\vspace{-0.05in}
\begin{tabular}{r|ccccc|cc|ccccc|cc}
\toprule
           & \multicolumn{7}{|c|}{General motion}                                                                                  & \multicolumn{7}{c}{Complex motion}                                                                                 \\
\hline
method         & d.d           & m.s.           & i2v-s.         & i2v-b.         & s.c.           & FVD            & FID            & d.d           & m.s.                 & i2v-s.         & i2v-b.         & s.c.           & FVD            & FID            \\
\hline
\hline
base finetuned & \textbf{0.90} & 0.986          & 0.974          & 0.979          & 0.921          & 383.4          & 42.68          & \underline{0.98}    & 0.985                & 0.971          & 0.976          & 0.913          & 437.0          & 52.01          \\
w/o g.m.       & \underline{0.85}    & 0.989          & 0.982          & 0.983          & 0.943          & 359.2          & 42.07          & 0.93          & 0.987                & 0.975          & 0.977          & 0.918          & 452.8          & 48.02          \\
w/o joint      & 0.73          & 0.989 & {0.983} & {0.984} & {0.946} & 354.5          & 40.42          & 0.96          & {0.988} & {0.978} & {0.979} & {0.926} & 411.8          & 47.2           \\
freeze dpt     & 0.77          & \textbf{0.991} & \underline{0.984}    & \underline{0.985}    & \textbf{0.956} & \textbf{336.0} & \underline{38.02}    & \underline{0.98}    & \textbf{0.990}       & \textbf{0.981} & \textbf{0.981} & \textbf{0.94}  & \textbf{382.3} & \underline{45.33}    \\
full           & 0.73          & \underline{0.990}    & \textbf{0.986} & \textbf{0.986} & \underline{0.953}    & \underline{336.1}    & \textbf{36.58} & \textbf{1.00} & 0.988                & \underline{0.980}    & \textbf{0.981} & \underline{0.928}    & \underline{394.2}    & \textbf{44.95} \\
\bottomrule
\end{tabular}
\label{table:videogen_ablation}
\vspace{-0.1in}
\end{table*}

\begin{figure*}[t]
    \centering
    \includegraphics[width=0.96\linewidth]{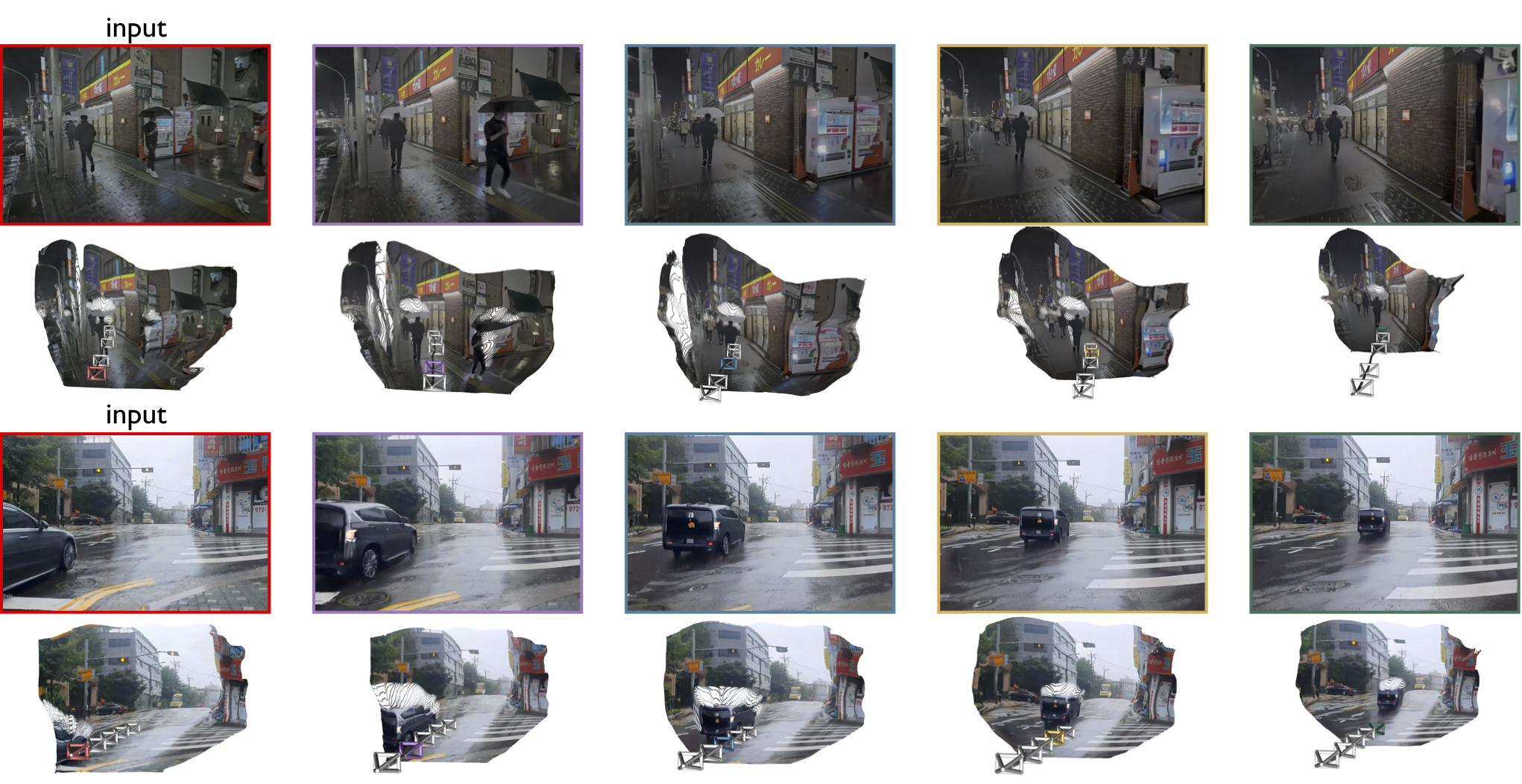}
    \caption{Qualitative 4D generation and geometry. For two in-the-wild inputs (left, red boxes), we show selected frames from our generated videos (top rows) alongside the corresponding \emph{dynamic point clouds} rendered from our pointmaps and camera trajectories (bottom rows). The persistent structure and consistent camera/object motion illustrate a single, stable 3D scene across time, evidencing strong geometric consistency in the underlying world state. See supplementary for additional examples.}
    \label{fig:4dgen_result}
\end{figure*}

\noindent\textbf{Ablation.}
We present an ablation study on video generation results in Table~\ref{table:videogen_ablation}. We show the effectiveness of two main designs in our method: (i) the geo-motion augmented latent for the DiT model; (ii) joint optimization of generation and 4D modeling with geometry and motion regularization. 
Removing the geo-motion augmented latent causes a clear drop in performance, especially on the \emph{complex} set. Applying joint training with regularization on the RGB-only model (``w/o g.m.") leads to even worse FVD results than simply finetuning the base model (``base finetuned") on the complex set. This demonstrates that our geo-motion latents are critical for modeling complex dynamics.
On the other hand, removing the joint training stage (``w/o joint") also hurts performance, which confirms that our regularization terms, applied via joint optimization, successfully align appearance and geometry, and improve motion quality.
Additionally, we present the results following the GeoVideo~\cite{bai2025geovideo} setup, where we freeze the temporal DPT heads (``freeze dpt") during joint training and apply only regularization terms. 

\section{Conclusion}

We have introduced \name, a feed-forward 4D video generator that natively couples appearance with object geometry and dynamics. By jointly emitting RGB, pointmaps, camera trajectories, and dense (scene and optical) flow, \name maintains a single persistent scene state, yielding consistent motion and stable geometry even for dynamic scenes. Trained in a mix of synthetic and real videos, \name achieves state-of-the-art 4D consistency while requiring no extra input at inference time, moving video generation one step closer to editable and agent-ready world models.

\noindent \textbf{Limitations and Future Work.} 
\name requires additional 4D supervision during training (e.g., camera, geometry, scene flow), which we currently obtain from synthetic data; 
Although \name introduces strategies to mitigate domain gaps, domain gaps still remain that constrain generalization to unusual motion and dynamics.
Also, since the temporal window is finite, failure modes appear under significant topology changes, heavy occlusions, and fast motions. We expect future work to reduce supervision by leveraging weak/self-supervised 4D signals from monocular videos, extend temporal context with streaming/causal diffusion and a persistent world state, and add controllable scene decomposition for more faithful long-horizon, interactive 4D generation.

{
    \small
    \bibliographystyle{ieeenat_fullname}
    \bibliography{main}

@String(CVPR= {IEEE Conf. Comput. Vis. Pattern Recog.})

@String(ICCV= {Int. Conf. Comput. Vis.})

@String(ICLR = {Int. Conf. Learn. Represent.})

@String(CVPR  = {CVPR})

@String(ICCV  = {ICCV})

@String(ICLR  = {ICLR})

@article{saharia2022photorealistic,
  title={Photorealistic text-to-image diffusion models with deep language understanding},
  author={Saharia, Chitwan and Chan, William and Saxena, Saurabh and Li, Lala and Whang, Jay and Denton, Emily L and Ghasemipour, Kamyar and Gontijo Lopes, Raphael and Karagol Ayan, Burcu and Salimans, Tim and others},
  journal={Advances in neural information processing systems},
  volume={35},
  pages={36479--36494},
  year={2022}
}

@inproceedings{hatamizadeh2024diffit,
  title={Diffit: Diffusion vision transformers for image generation},
  author={Hatamizadeh, Ali and Song, Jiaming and Liu, Guilin and Kautz, Jan and Vahdat, Arash},
  booktitle={European Conference on Computer Vision},
  pages={37--55},
  year={2024},
  organization={Springer}
}

@article{guo2023animatediff,
  title={Animatediff: Animate your personalized text-to-image diffusion models without specific tuning},
  author={Guo, Yuwei and Yang, Ceyuan and Rao, Anyi and Liang, Zhengyang and Wang, Yaohui and Qiao, Yu and Agrawala, Maneesh and Lin, Dahua and Dai, Bo},
  journal={arXiv preprint arXiv:2307.04725},
  year={2023}
}

@article{ho2022video,
  title={Video diffusion models},
  author={Ho, Jonathan and Salimans, Tim and Gritsenko, Alexey and Chan, William and Norouzi, Mohammad and Fleet, David J},
  journal={Advances in neural information processing systems},
  volume={35},
  pages={8633--8646},
  year={2022}
}

@article{singer2022make,
  title={Make-a-video: Text-to-video generation without text-video data},
  author={Singer, Uriel and Polyak, Adam and Hayes, Thomas and Yin, Xi and An, Jie and Zhang, Songyang and Hu, Qiyuan and Yang, Harry and Ashual, Oron and Gafni, Oran and others},
  journal={arXiv preprint arXiv:2209.14792},
  year={2022}
}

@inproceedings{wu2023tune,
  title={Tune-a-video: One-shot tuning of image diffusion models for text-to-video generation},
  author={Wu, Jay Zhangjie and Ge, Yixiao and Wang, Xintao and Lei, Stan Weixian and Gu, Yuchao and Shi, Yufei and Hsu, Wynne and Shan, Ying and Qie, Xiaohu and Shou, Mike Zheng},
  booktitle={Proceedings of the IEEE/CVF international conference on computer vision},
  pages={7623--7633},
  year={2023}
}

@inproceedings{rombach2022high,
  title={High-resolution image synthesis with latent diffusion models},
  author={Rombach, Robin and Blattmann, Andreas and Lorenz, Dominik and Esser, Patrick and Ommer, Bj{\"o}rn},
  booktitle={Proceedings of the IEEE/CVF conference on computer vision and pattern recognition},
  pages={10684--10695},
  year={2022}
}

@article{vahdat2021score,
  title={Score-based generative modeling in latent space},
  author={Vahdat, Arash and Kreis, Karsten and Kautz, Jan},
  journal={Advances in neural information processing systems},
  volume={34},
  pages={11287--11302},
  year={2021}
}

@article{kingma2013auto,
  title={Auto-encoding variational bayes},
  author={Kingma, Diederik P and Welling, Max},
  journal={arXiv preprint arXiv:1312.6114},
  year={2013}
}

@article{blattmann2023stable,
  title={Stable video diffusion: Scaling latent video diffusion models to large datasets},
  author={Blattmann, Andreas and Dockhorn, Tim and Kulal, Sumith and Mendelevitch, Daniel and Kilian, Maciej and Lorenz, Dominik and Levi, Yam and English, Zion and Voleti, Vikram and Letts, Adam and others},
  journal={arXiv preprint arXiv:2311.15127},
  year={2023}
}

@article{chen2023videocrafter1,
  title={Videocrafter1: Open diffusion models for high-quality video generation},
  author={Chen, Haoxin and Xia, Menghan and He, Yingqing and Zhang, Yong and Cun, Xiaodong and Yang, Shaoshu and Xing, Jinbo and Liu, Yaofang and Chen, Qifeng and Wang, Xintao and others},
  journal={arXiv preprint arXiv:2310.19512},
  year={2023}
}

@article{he2022latent,
  title={Latent video diffusion models for high-fidelity long video generation},
  author={He, Yingqing and Yang, Tianyu and Zhang, Yong and Shan, Ying and Chen, Qifeng},
  journal={arXiv preprint arXiv:2211.13221},
  year={2022}
}

@article{ho2022imagen,
  title={Imagen video: High definition video generation with diffusion models},
  author={Ho, Jonathan and Chan, William and Saharia, Chitwan and Whang, Jay and Gao, Ruiqi and Gritsenko, Alexey and Kingma, Diederik P and Poole, Ben and Norouzi, Mohammad and Fleet, David J and others},
  journal={arXiv preprint arXiv:2210.02303},
  year={2022}
}

@article{wang2025lavie,
  title={Lavie: High-quality video generation with cascaded latent diffusion models},
  author={Wang, Yaohui and Chen, Xinyuan and Ma, Xin and Zhou, Shangchen and Huang, Ziqi and Wang, Yi and Yang, Ceyuan and He, Yinan and Yu, Jiashuo and Yang, Peiqing and others},
  journal={International Journal of Computer Vision},
  volume={133},
  number={5},
  pages={3059--3078},
  year={2025},
  publisher={Springer}
}

@article{zhang2023i2vgen,
  title={I2vgen-xl: High-quality image-to-video synthesis via cascaded diffusion models},
  author={Zhang, Shiwei and Wang, Jiayu and Zhang, Yingya and Zhao, Kang and Yuan, Hangjie and Qin, Zhiwu and Wang, Xiang and Zhao, Deli and Zhou, Jingren},
  journal={arXiv preprint arXiv:2311.04145},
  year={2023}
}

@article{kong2024hunyuanvideo,
  title={Hunyuanvideo: A systematic framework for large video generative models},
  author={Kong, Weijie and Tian, Qi and Zhang, Zijian and Min, Rox and Dai, Zuozhuo and Zhou, Jin and Xiong, Jiangfeng and Li, Xin and Wu, Bo and Zhang, Jianwei and others},
  journal={arXiv preprint arXiv:2412.03603},
  year={2024}
}

@article{yang2024cogvideox,
  title={Cogvideox: Text-to-video diffusion models with an expert transformer},
  author={Yang, Zhuoyi and Teng, Jiayan and Zheng, Wendi and Ding, Ming and Huang, Shiyu and Xu, Jiazheng and Yang, Yuanming and Hong, Wenyi and Zhang, Xiaohan and Feng, Guanyu and others},
  journal={arXiv preprint arXiv:2408.06072},
  year={2024}
}

@article{hong2022cogvideo,
  title={Cogvideo: Large-scale pretraining for text-to-video generation via transformers},
  author={Hong, Wenyi and Ding, Ming and Zheng, Wendi and Liu, Xinghan and Tang, Jie},
  journal={arXiv preprint arXiv:2205.15868},
  year={2022}
}

@article{wan2025wan,
  title={Wan: Open and advanced large-scale video generative models},
  author={Wan, Team and Wang, Ang and Ai, Baole and Wen, Bin and Mao, Chaojie and Xie, Chen-Wei and Chen, Di and Yu, Feiwu and Zhao, Haiming and Yang, Jianxiao and others},
  journal={arXiv preprint arXiv:2503.20314},
  year={2025}
}

@inproceedings{peebles2023scalable,
  title={Scalable diffusion models with transformers},
  author={Peebles, William and Xie, Saining},
  booktitle={Proceedings of the IEEE/CVF international conference on computer vision},
  pages={4195--4205},
  year={2023}
}

@inproceedings{zhang2025world,
  title={World-consistent video diffusion with explicit 3d modeling},
  author={Zhang, Qihang and Zhai, Shuangfei and Martin, Miguel Angel Bautista and Miao, Kevin and Toshev, Alexander and Susskind, Joshua and Gu, Jiatao},
  booktitle={Proceedings of the Computer Vision and Pattern Recognition Conference},
  pages={21685--21695},
  year={2025}
}

@article{huang2025voyager,
  title={Voyager: Long-Range and World-Consistent Video Diffusion for Explorable 3D Scene Generation},
  author={Huang, Tianyu and Zheng, Wangguandong and Wang, Tengfei and Liu, Yuhao and Wang, Zhenwei and Wu, Junta and Jiang, Jie and Li, Hui and Lau, Rynson WH and Zuo, Wangmeng and others},
  journal={arXiv preprint arXiv:2506.04225},
  year={2025}
}

@article{dai2025fantasyworld,
  title={FantasyWorld: Geometry-Consistent World Modeling via Unified Video and 3D Prediction},
  author={Dai, Yixiang and Jiang, Fan and Wang, Chiyu and Xu, Mu and Qi, Yonggang},
  journal={arXiv preprint arXiv:2509.21657},
  year={2025}
}

@article{xi2025omnivdiff,
  title={Omnivdiff: Omni controllable video diffusion for generation and understanding},
  author={Xi, Dianbing and Wang, Jiepeng and Liang, Yuanzhi and Qiu, Xi and Huo, Yuchi and Wang, Rui and Zhang, Chi and Li, Xuelong},
  journal={arXiv preprint arXiv:2504.10825},
  year={2025}
}

@article{wu2025geometry,
  title={Geometry forcing: Marrying video diffusion and 3d representation for consistent world modeling},
  author={Wu, Haoyu and Wu, Diankun and He, Tianyu and Guo, Junliang and Ye, Yang and Duan, Yueqi and Bian, Jiang},
  journal={arXiv preprint arXiv:2507.07982},
  year={2025}
}

@article{chefer2025videojam,
  title={Videojam: Joint appearance-motion representations for enhanced motion generation in video models},
  author={Chefer, Hila and Singer, Uriel and Zohar, Amit and Kirstain, Yuval and Polyak, Adam and Taigman, Yaniv and Wolf, Lior and Sheynin, Shelly},
  journal={arXiv preprint arXiv:2502.02492},
  year={2025}
}

@inproceedings{jeong2025track4gen,
  title={Track4gen: Teaching video diffusion models to track points improves video generation},
  author={Jeong, Hyeonho and Huang, Chun-Hao P and Ye, Jong Chul and Mitra, Niloy J and Ceylan, Duygu},
  booktitle={Proceedings of the Computer Vision and Pattern Recognition Conference},
  pages={7276--7287},
  year={2025}
}

@inproceedings{shi2024motion,
  title={Motion-i2v: Consistent and controllable image-to-video generation with explicit motion modeling},
  author={Shi, Xiaoyu and Huang, Zhaoyang and Wang, Fu-Yun and Bian, Weikang and Li, Dasong and Zhang, Yi and Zhang, Manyuan and Cheung, Ka Chun and See, Simon and Qin, Hongwei and others},
  booktitle={ACM SIGGRAPH 2024 Conference Papers},
  pages={1--11},
  year={2024}
}

@inproceedings{jin2025flovd,
  title={Flovd: Optical flow meets video diffusion model for enhanced camera-controlled video synthesis},
  author={Jin, Wonjoon and Dai, Qi and Luo, Chong and Baek, Seung-Hwan and Cho, Sunghyun},
  booktitle={Proceedings of the Computer Vision and Pattern Recognition Conference},
  pages={2040--2049},
  year={2025}
}

@inproceedings{nam2025optical,
  title={Optical-flow guided prompt optimization for coherent video generation},
  author={Nam, Hyelin and Kim, Jaemin and Lee, Dohun and Ye, Jong Chul},
  booktitle={Proceedings of the Computer Vision and Pattern Recognition Conference},
  pages={7837--7846},
  year={2025}
}

@inproceedings{wang2025motif,
  title={MotiF: Making Text Count in Image Animation with Motion Focal Loss},
  author={Wang, Shijie and Azadi, Samaneh and Girdhar, Rohit and Rambhatla, Saketh and Sun, Chen and Yin, Xi},
  booktitle={Proceedings of the Computer Vision and Pattern Recognition Conference},
  pages={7773--7783},
  year={2025}
}

@inproceedings{wang2024dust3r,
  title={Dust3r: Geometric 3d vision made easy},
  author={Wang, Shuzhe and Leroy, Vincent and Cabon, Yohann and Chidlovskii, Boris and Revaud, Jerome},
  booktitle={Proceedings of the IEEE/CVF Conference on Computer Vision and Pattern Recognition},
  pages={20697--20709},
  year={2024}
}

@inproceedings{wang2025vggt,
  title={Vggt: Visual geometry grounded transformer},
  author={Wang, Jianyuan and Chen, Minghao and Karaev, Nikita and Vedaldi, Andrea and Rupprecht, Christian and Novotny, David},
  booktitle={Proceedings of the Computer Vision and Pattern Recognition Conference},
  pages={5294--5306},
  year={2025}
}

@article{zhang2024monst3r,
  title={Monst3r: A simple approach for estimating geometry in the presence of motion},
  author={Zhang, Junyi and Herrmann, Charles and Hur, Junhwa and Jampani, Varun and Darrell, Trevor and Cole, Forrester and Sun, Deqing and Yang, Ming-Hsuan},
  journal={arXiv preprint arXiv:2410.03825},
  year={2024}
}

@article{chen2025easi3r,
  title={Easi3r: Estimating disentangled motion from dust3r without training},
  author={Chen, Xingyu and Chen, Yue and Xiu, Yuliang and Geiger, Andreas and Chen, Anpei},
  journal={arXiv preprint arXiv:2503.24391},
  year={2025}
}

@misc{wang2025continuous3dperceptionmodel,
      title={Continuous 3D Perception Model with Persistent State}, 
      author={Qianqian Wang and Yifei Zhang and Aleksander Holynski and Alexei A. Efros and Angjoo Kanazawa},
      year={2025},
      eprint={2501.12387},
      archivePrefix={arXiv},
      primaryClass={cs.CV},
      url={https://arxiv.org/abs/2501.12387}, 
}

@INPROCEEDINGS{11092390,
  author={Jin, Linyi and Tucker, Richard and Li, Zhengqi and Fouhey, David and Snavely, Noah and Hołyński, Aleksander},
  booktitle={2025 IEEE/CVF Conference on Computer Vision and Pattern Recognition (CVPR)}, 
  title={Stereo4D: Learning How Things Move in 3D from Internet Stereo Videos}, 
  year={2025},
  volume={},
  number={},
  pages={10497-10509},
  doi={10.1109/CVPR52734.2025.00982}}

@inproceedings{huang2025vipe,
        title={ViPE: Video Pose Engine for 3D Geometric Perception},
        author={Huang, Jiahui and Zhou, Qunjie and Rabeti, Hesam and Korovko, Aleksandr and Ling, Huan and Ren, Xuanchi and Shen, Tianchang and Gao, Jun and Slepichev, Dmitry and Lin, Chen-Hsuan and Ren, Jiawei and Xie, Kevin and Biswas, Joydeep and Leal-Taixe, Laura and Fidler, Sanja},
        booktitle={NVIDIA Research Whitepapers},
        year={2025}
    }

@article{badki2025l4p,
  author    = {Badki, Abhishek and Su, Hang and Wen, Bowen and Gallo, Orazio},
  title     = {{L4P}: Towards Unified {L}ow-Level {4D} Vision Perception},
  journal   = arxiv,
  year      = {2025},
}

@misc{jiang2025geo4d,
      title={Geo4D: Leveraging Video Generators for Geometric 4D Scene Reconstruction}, 
      author={Zeren Jiang and Chuanxia Zheng and Iro Laina and Diane Larlus and Andrea Vedaldi},
      year={2025},
      eprint={2504.07961},
      archivePrefix={arXiv},
      primaryClass={cs.CV},
      url={https://arxiv.org/abs/2504.07961}, 
}

@article{xu2025geometrycrafter,
  title={GeometryCrafter: Consistent Geometry Estimation for Open-world Videos with Diffusion Priors},
  author={Xu, Tian-Xing and Gao, Xiangjun and Hu, Wenbo and Li, Xiaoyu and Zhang, Song-Hai and Shan, Ying},
  journal={arXiv preprint arXiv:2504.01016},
  year={2025}
}

@article{poole2022dreamfusion,
  author = {Poole, Ben and Jain, Ajay and Barron, Jonathan T. and Mildenhall, Ben},
  title = {DreamFusion: Text-to-3D using 2D Diffusion},
  journal = {arXiv},
  year = {2022},
}

@article{pumarola2020d,
  title={D-NeRF: Neural Radiance Fields for Dynamic Scenes},
  author={Pumarola, Albert and Corona, Enric and Pons-Moll, Gerard and Moreno-Noguer, Francesc},
  journal={arXiv preprint arXiv:2011.13961},
  year={2020}
}

@InProceedings{Wu_2024_CVPR,
    author    = {Wu, Guanjun and Yi, Taoran and Fang, Jiemin and Xie, Lingxi and Zhang, Xiaopeng and Wei, Wei and Liu, Wenyu and Tian, Qi and Wang, Xinggang},
    title     = {4D Gaussian Splatting for Real-Time Dynamic Scene Rendering},
    booktitle = {Proceedings of the IEEE/CVF Conference on Computer Vision and Pattern Recognition (CVPR)},
    month     = {June},
    year      = {2024},
    pages     = {20310-20320}
}

@article{singer2023text4d,
  author = {Singer, Uriel and Sheynin, Shelly and Polyak, Adam and Ashual, Oron and
           Makarov, Iurii and Kokkinos, Filippos and Goyal, Naman and Vedaldi, Andrea and
           Parikh, Devi and Johnson, Justin and Taigman, Yaniv},
  title = {Text-To-4D Dynamic Scene Generation},
  journal = {arXiv:2301.11280},
  year = {2023},
}

@article{ren2023dreamgaussian4d,
  title={DreamGaussian4D: Generative 4D Gaussian Splatting},
  author={Ren, Jiawei and Pan, Liang and Tang, Jiaxiang and Zhang, Chi and Cao, Ang and Zeng, Gang and Liu, Ziwei},
  journal={arXiv preprint arXiv:2312.17142},
  year={2023}
}

@article{bah20244dfy,
  author = {Bahmani, Sherwin and Skorokhodov, Ivan and Rong, Victor and Wetzstein, Gordon and Guibas, Leonidas and Wonka, Peter and Tulyakov, Sergey and Park, Jeong Joon and Tagliasacchi, Andrea and Lindell, David B.},
  title = {4D-fy: Text-to-4D Generation Using Hybrid Score Distillation Sampling},
  journal = {IEEE Conference on Computer Vision and Pattern Recognition ({CVPR})},
  year = {2024},
}

@article{pan2024efficient4d,
  title={Efficient4D: Fast Dynamic 3D Object Generation from a Single-view Video},
  author={Pan, Zijie and Yang, Zeyu and Zhu, Xiatian and Zhang, Li},
  journal={arXiv preprint arXiv 2401.08742},
  year={2024}
}

@inproceedings{zhao2024genxd,
  author={Zhao, Yuyang and Lin, Chung-Ching and Lin, Kevin and Yan, Zhiwen and Li, Linjie and Yang, Zhengyuan and Wang, Jianfeng and Lee, Gim Hee and Wang, Lijuan},
  title={GenXD: Generating Any 3D and 4D Scenes},
  booktitle={ICLR},
  year={2025}
}

@inproceedings{sun2024dimensionx,
  title={DimensionX: Create Any 3D and 4D Scenes from a Single Image with Controllable Video Diffusion},
  author={Sun, Wenqiang and Chen, Shuo and Liu, Fangfu and Chen, Zilong and Duan, Yueqi and Zhang, Jun and Wang, Yikai},
  booktitle={International Conference on Computer Vision (ICCV)},
  year={2025}
}

@inproceedings{ren2025gen3c,
    title={GEN3C: 3D-Informed World-Consistent Video Generation with Precise Camera Control},
    author={Ren, Xuanchi and Shen, Tianchang and Huang, Jiahui and Ling, Huan and 
        Lu, Yifan and Nimier-David, Merlin and M\"uller, Thomas and Keller, Alexander and 
        Fidler, Sanja and Gao, Jun},
    booktitle={Proceedings of the IEEE/CVF Conference on Computer Vision and Pattern Recognition},
    year={2025}
}

@inproceedings{ren2024l4gm,
    title={L4GM: Large 4D Gaussian Reconstruction Model}, 
    author={Jiawei Ren and Kevin Xie and Ashkan Mirzaei and Hanxue Liang and Xiaohui Zeng and Karsten Kreis and Ziwei Liu and Antonio Torralba and Sanja Fidler and Seung Wook Kim and Huan Ling},
    booktitle={Proceedings of Neural Information Processing Systems(NeurIPS)},
    month = {Dec},
    year={2024}
}

@article{tang2024lgm,
  title={LGM: Large Multi-View Gaussian Model for High-Resolution 3D Content Creation},
  author={Tang, Jiaxiang and Chen, Zhaoxi and Chen, Xiaokang and Wang, Tengfei and Zeng, Gang and Liu, Ziwei},
  journal={arXiv preprint arXiv:2402.05054},
  year={2024}
}

@article{zhen2025tesseract,
  title={TesserAct: Learning 4D Embodied World Models}, 
  author={Haoyu Zhen and Qiao Sun and Hongxin Zhang and Junyan Li and Siyuan Zhou and Yilun Du and Chuang Gan},
  year={2025},
  eprint={2504.20995},
  archivePrefix={arXiv},
  primaryClass={cs.CV},
  url={https://arxiv.org/abs/2504.20995}, 
}

@article{chen20254dnex,
    title={4DNeX: Feed-Forward 4D Generative Modeling Made Easy},
    author={Chen, Zhaoxi and Liu, Tianqi and Zhuo, Long and Ren, Jiawei and Tao, Zeng and Zhu, He and Hong, Fangzhou and Pan, Liang and Liu, Ziwei},
    journal={arXiv preprint arXiv:2508.13154},
    year={2025}
}

@article{han2025d,
  title={D\^{} 2USt3R: Enhancing 3D Reconstruction with 4D Pointmaps for Dynamic Scenes},
  author={Han, Jisang and An, Honggyu and Jung, Jaewoo and Narihira, Takuya and Seo, Junyoung and Fukuda, Kazumi and Kim, Chaehyun and Hong, Sunghwan and Mitsufuji, Yuki and Kim, Seungryong},
  journal={arXiv preprint arXiv:2504.06264},
  year={2025}
}

@article{zhuo2025streaming,
  title={Streaming 4d visual geometry transformer},
  author={Zhuo, Dong and Zheng, Wenzhao and Guo, Jiahe and Wu, Yuqi and Zhou, Jie and Lu, Jiwen},
  journal={arXiv preprint arXiv:2507.11539},
  year={2025}
}

@inproceedings{xing2024dynamicrafter,
  title={Dynamicrafter: Animating open-domain images with video diffusion priors},
  author={Xing, Jinbo and Xia, Menghan and Zhang, Yong and Chen, Haoxin and Yu, Wangbo and Liu, Hanyuan and Liu, Gongye and Wang, Xintao and Shan, Ying and Wong, Tien-Tsin},
  booktitle={European Conference on Computer Vision},
  pages={399--417},
  year={2024},
  organization={Springer}
}

@inproceedings{wang2023videomae,
  title={Videomae v2: Scaling video masked autoencoders with dual masking},
  author={Wang, Limin and Huang, Bingkun and Zhao, Zhiyu and Tong, Zhan and He, Yinan and Wang, Yi and Wang, Yali and Qiao, Yu},
  booktitle={Proceedings of the IEEE/CVF conference on computer vision and pattern recognition},
  pages={14549--14560},
  year={2023}
}

@misc{liu2023zero1to3,
      title={Zero-1-to-3: Zero-shot One Image to 3D Object}, 
      author={Ruoshi Liu and Rundi Wu and Basile Van Hoorick and Pavel Tokmakov and Sergey Zakharov and Carl Vondrick},
      year={2023},
      eprint={2303.11328},
      archivePrefix={arXiv},
      primaryClass={cs.CV}
}

@article{shi2023MVDream,
  author = {Shi, Yichun and Wang, Peng and Ye, Jianglong and Mai, Long and Li, Kejie and Yang, Xiao},
  title = {MVDream: Multi-view Diffusion for 3D Generation},
  journal = {arXiv:2308.16512},
  year = {2023},
}

@inproceedings{wu2025cat4d,
  title={Cat4d: Create anything in 4d with multi-view video diffusion models},
  author={Wu, Rundi and Gao, Ruiqi and Poole, Ben and Trevithick, Alex and Zheng, Changxi and Barron, Jonathan T and Holynski, Aleksander},
  booktitle={Proceedings of the Computer Vision and Pattern Recognition Conference},
  pages={26057--26068},
  year={2025}
}

@inproceedings{zheng2023pointodyssey,
  title={Pointodyssey: A large-scale synthetic dataset for long-term point tracking},
  author={Zheng, Yang and Harley, Adam W and Shen, Bokui and Wetzstein, Gordon and Guibas, Leonidas J},
  booktitle={Proceedings of the IEEE/CVF International Conference on Computer Vision},
  pages={19855--19865},
  year={2023}
}

@inproceedings{black2023bedlam,
  title={Bedlam: A synthetic dataset of bodies exhibiting detailed lifelike animated motion},
  author={Black, Michael J and Patel, Priyanka and Tesch, Joachim and Yang, Jinlong},
  booktitle={Proceedings of the IEEE/CVF Conference on Computer Vision and Pattern Recognition},
  pages={8726--8737},
  year={2023}
}

@article{karaev2023dynamicstereo,
  title={DynamicStereo: Consistent Dynamic Depth from Stereo Videos},
  author={Nikita Karaev and Ignacio Rocco and Benjamin Graham and Natalia Neverova and Andrea Vedaldi and Christian Rupprecht},
  journal={CVPR},
  year={2023}
}

@misc{zhou2025omniworld,
      title={OmniWorld: A Multi-Domain and Multi-Modal Dataset for 4D World Modeling}, 
      author={Yang Zhou and Yifan Wang and Jianjun Zhou and Wenzheng Chang and Haoyu Guo and Zizun Li and Kaijing Ma and Xinyue Li and Yating Wang and Haoyi Zhu and Mingyu Liu and Dingning Liu and Jiange Yang and Zhoujie Fu and Junyi Chen and Chunhua Shen and Jiangmiao Pang and Kaipeng Zhang and Tong He},
      year={2025},
      eprint={2509.12201},
      archivePrefix={arXiv},
      primaryClass={cs.CV},
      url={https://arxiv.org/abs/2509.12201}, 
}

@misc{wang2025spatialvidlargescalevideodataset,
      title={SpatialVID: A Large-Scale Video Dataset with Spatial Annotations}, 
      author={Jiahao Wang and Yufeng Yuan and Rujie Zheng and Youtian Lin and Jian Gao and Lin-Zhuo Chen and Yajie Bao and Yi Zhang and Chang Zeng and Yanxi Zhou and Xiaoxiao Long and Hao Zhu and Zhaoxiang Zhang and Xun Cao and Yao Yao},
      year={2025},
      eprint={2509.09676},
      archivePrefix={arXiv},
      primaryClass={cs.CV},
      url={https://arxiv.org/abs/2509.09676}, 
}

@article{zheng2024open,
  title={Open-sora: Democratizing efficient video production for all},
  author={Zheng, Zangwei and Peng, Xiangyu and Yang, Tianji and Shen, Chenhui and Li, Shenggui and Liu, Hongxin and Zhou, Yukun and Li, Tianyi and You, Yang},
  journal={arXiv preprint arXiv:2412.20404},
  year={2024}
}

@inproceedings{hu2025depthcrafter,
  title={Depthcrafter: Generating consistent long depth sequences for open-world videos},
  author={Hu, Wenbo and Gao, Xiangjun and Li, Xiaoyu and Zhao, Sijie and Cun, Xiaodong and Zhang, Yong and Quan, Long and Shan, Ying},
  booktitle={Proceedings of the Computer Vision and Pattern Recognition Conference},
  pages={2005--2015},
  year={2025}
}

@inproceedings{wang2024sea,
  title={Sea-raft: Simple, efficient, accurate raft for optical flow},
  author={Wang, Yihan and Lipson, Lahav and Deng, Jia},
  booktitle={European Conference on Computer Vision},
  pages={36--54},
  year={2024},
  organization={Springer}
}

@inproceedings{chen2024panda,
  title={Panda-70m: Captioning 70m videos with multiple cross-modality teachers},
  author={Chen, Tsai-Shien and Siarohin, Aliaksandr and Menapace, Willi and Deyneka, Ekaterina and Chao, Hsiang-wei and Jeon, Byung Eun and Fang, Yuwei and Lee, Hsin-Ying and Ren, Jian and Yang, Ming-Hsuan and others},
  booktitle={Proceedings of the IEEE/CVF Conference on Computer Vision and Pattern Recognition},
  pages={13320--13331},
  year={2024}
}

@article{fan2025vchitect,
  title={Vchitect-2.0: Parallel transformer for scaling up video diffusion models},
  author={Fan, Weichen and Si, Chenyang and Song, Junhao and Yang, Zhenyu and He, Yinan and Zhuo, Long and Huang, Ziqi and Dong, Ziyue and He, Jingwen and Pan, Dongwei and others},
  journal={arXiv preprint arXiv:2501.08453},
  year={2025}
}

@inproceedings{li2025megasam,
  title={MegaSaM: Accurate, fast and robust structure and motion from casual dynamic videos},
  author={Li, Zhengqi and Tucker, Richard and Cole, Forrester and Wang, Qianqian and Jin, Linyi and Ye, Vickie and Kanazawa, Angjoo and Holynski, Aleksander and Snavely, Noah},
  booktitle={Proceedings of the Computer Vision and Pattern Recognition Conference},
  pages={10486--10496},
  year={2025}
}

@inproceedings{liang2025zero,
  title={Zero-shot monocular scene flow estimation in the wild},
  author={Liang, Yiqing and Badki, Abhishek and Su, Hang and Tompkin, James and Gallo, Orazio},
  booktitle={Proceedings of the Computer Vision and Pattern Recognition Conference},
  pages={21031--21044},
  year={2025}
}

@inproceedings{ranftl2021vision,
  title={Vision transformers for dense prediction},
  author={Ranftl, Ren{\'e} and Bochkovskiy, Alexey and Koltun, Vladlen},
  booktitle={Proceedings of the IEEE/CVF international conference on computer vision},
  pages={12179--12188},
  year={2021}
}

@inproceedings{
bai2025geovideo,
title={GeoVideo: Introducing Geometric Regularization into Video Generation Model},
author={Yunpeng Bai and Shaoheng Fang and Chaohui Yu and Fan Wang and Qixing Huang},
booktitle={The Thirty-ninth Annual Conference on Neural Information Processing Systems},
year={2025},
url={https://openreview.net/forum?id=be6u40gq0Y}
}

@article{loshchilov2017fixing,
  title={Fixing weight decay regularization in adam},
  author={Loshchilov, Ilya and Hutter, Frank and others},
  journal={arXiv preprint arXiv:1711.05101},
  volume={5},
  number={5},
  pages={5},
  year={2017}
}

@inproceedings{huang2024vbench,
  title={Vbench: Comprehensive benchmark suite for video generative models},
  author={Huang, Ziqi and He, Yinan and Yu, Jiashuo and Zhang, Fan and Si, Chenyang and Jiang, Yuming and Zhang, Yuanhan and Wu, Tianxing and Jin, Qingyang and Chanpaisit, Nattapol and others},
  booktitle={Proceedings of the IEEE/CVF Conference on Computer Vision and Pattern Recognition},
  pages={21807--21818},
  year={2024}
}

@article{unterthiner2018towards,
  title={Towards accurate generative models of video: A new metric \& challenges},
  author={Unterthiner, Thomas and Van Steenkiste, Sjoerd and Kurach, Karol and Marinier, Raphael and Michalski, Marcin and Gelly, Sylvain},
  journal={arXiv preprint arXiv:1812.01717},
  year={2018}
}

@article{heusel2017gans,
  title={Gans trained by a two time-scale update rule converge to a local nash equilibrium},
  author={Heusel, Martin and Ramsauer, Hubert and Unterthiner, Thomas and Nessler, Bernhard and Hochreiter, Sepp},
  journal={Advances in neural information processing systems},
  volume={30},
  year={2017}
}

@inproceedings{ke2024repurposing,
  title={Repurposing diffusion-based image generators for monocular depth estimation},
  author={Ke, Bingxin and Obukhov, Anton and Huang, Shengyu and Metzger, Nando and Daudt, Rodrigo Caye and Schindler, Konrad},
  booktitle={Proceedings of the IEEE/CVF conference on computer vision and pattern recognition},
  pages={9492--9502},
  year={2024}
}
}


\end{document}